\documentclass{article}

\usepackage{times}
\usepackage{graphicx} 
\usepackage{subfigure} 

\usepackage{natbib}

\usepackage{algorithm}
\usepackage{algorithmic}

\usepackage{hyperref}

\usepackage[accepted]{icml2015}

\usepackage{dsfont}
\usepackage{wrapfig}
\usepackage[colorinlistoftodos]{todonotes}
\usepackage{amssymb, amsmath}

\usepackage{color}
\usepackage{multirow}

\newcommand{\R}{\mathds{R}}
\renewcommand{\vec}{\boldsymbol}
\newcommand{\mat}{\boldsymbol}
\newcommand{\E}{\mathds{E}}
\newcommand{\var}{\mathrm{var}}

\newcommand{\T}{^T}
\newcommand{\inv}{^{-1}}

\newcommand{\gauss}[2]{\mathcal N(#1,#2)}

\newcommand{\figspace}{\vspace{-3mm}}
\newcommand{\idx}[1]{{(#1)}}

\definecolor{darkred}{RGB}{180,0,0}
\newcommand{\red}[1]{\bf{\color{darkred}{#1}}}
\definecolor{blue}{RGB}{0,0,180}
\newcommand{\blue}[1]{\bf{\color{blue}{#1}}}

\icmltitlerunning{Distributed Gaussian Processes}

\begin{document} 
\twocolumn[ \icmltitle{Distributed Gaussian Processes}

\icmlauthor{Marc Peter Deisenroth}{m.deisenroth@imperial.ac.uk}
\icmladdress{Department of Computing, Imperial College London, United Kingdom}
\icmlauthor{Jun Wei Ng}{junwei.ng10@alumni.imperial.ac.uk}
\icmladdress{Department of Computing, Imperial College London, United Kingdom}
\icmlkeywords{Gaussian processes, distributed computation, Bayesian
  inference, Bayesian Committee Machine, Big Data}

\vskip 0.3in
]

\begin{abstract} 
To scale Gaussian processes (GPs) to large data sets we introduce the robust Bayesian Committee Machine (rBCM), a practical and scalable product-of-experts model for large-scale distributed GP regression. Unlike state-of-the-art sparse GP approximations, the rBCM is conceptually simple and does not rely on inducing or variational parameters. The key idea is to recursively distribute computations to independent computational units and, subsequently, recombine them to form an overall result. Efficient closed-form inference allows for straightforward parallelisation and distributed computations with a small memory footprint. The rBCM is independent of the computational graph and can be used on heterogeneous computing infrastructures, ranging from laptops to clusters. With sufficient computing resources our distributed GP model can handle arbitrarily large data sets. 
\end{abstract}

\section{Introduction}
 
Gaussian processes (GPs)~\cite{Rasmussen2006} are the method of choice for probabilistic nonlinear regression: Their non-parametric nature allows for flexible modelling without specifying low-level assumptions (e.g., the degree of a polynomial) in advance. Inference can be performed in a principled way simply by applying Bayes' theorem. GPs have had substantial impact in geostatistics~\cite{Cressie1993}, optimisation~\cite{Jones1998,Brochu2009}, data visualisation~\cite{Lawrence2005}, robotics and reinforcement learning~\cite{Deisenroth2015}, spatio-temporal modelling~\cite{Luttinen2012}, and active learning~\cite{Krause2008}. A strength of GPs is that they are a fairly reliable black-box function approximator, i.e., they produce reasonable predictions without manual parameter tuning.
However, their inherent weakness is that they scale poorly with the size $N$ of the data set: Training and predicting scale in $\mathcal{O}(N^3)$ and $\mathcal{O}(N^2)$, respectively, which practically limits GPs to data sets of size $\mathcal O(10^4)$.

For large data sets (e.g., $N>10^4$) sparse approximations are often used~\cite{Williams2001, Quinonero-Candela2005, Hensman2013, Titsias2009, Lazaro-Gredilla2010, Shen2006, Gal2014}. Typically, they lower the computational burden by implicitly (or explicitly) using a subset of the data, which scales (sparse) GPs to training set sizes $N\in\mathcal O(10^6)$. 
%
Recently, \citet{Gal2014} proposed an approach that scales variational sparse GPs~\cite{Titsias2009} further by exploiting distributed computations. In particular, they derive an exact re-parameterisation of the variational sparse GP by \citet{Titsias2009} to update the variational parameters independently on different computing nodes. However, even with sparse approximations it is inconceivable to apply GPs to training set sizes of $N\geq\mathcal O(10^7)$.

An alternative to sparse approximations is to distribute the computations by using independent local ``expert'' models, which operate on subsets of the data. These local models typically require stationary kernels for a notion of ``distance'' and ``locality''.
\citet{Shen2006} used KD-trees to recursively partition the data space into a multi-resolution tree data structure, which scale GPs to $\mathcal O(10^4)$ training points. However, no solutions for variance predictions are provided, and the approach is limited to stationary kernels.
Along the lines of exploiting locality, mixture-of-experts (MoE) models~\cite{Jacobs1991} have been applied to GP regression~\cite{Rasmussen2002,Meeds2006,Yuan2009}. However, these models have not primarily been used to speed up GP regression, but rather to increase the expressiveness of the model, i.e., allowing for heteroscedasticity and non-stationarity. Each local model possesses its own set of hyper-parameters to be optimised. Predictions are made by collecting the predictions of all local expert models, and weighting them using the responsibilities assigned by the gating network. Closed-form inference in these models is intractable, and approximations typically require MCMC. \citet{Nguyen2014} sidestep MCMC inference and speed the GP-MoE model up by (i) fixing the number of GP experts, (ii) combining it with the pseudo-input sparse approximation by \citet{Snelson2006}. This approach assigns data points to expert probabilistically using proximity information provided by stationary kernels and scales GPs to $\mathcal O(10^5)$ data points.

Product-of-GP-experts models (PoEs) sidestep the weight assignment problem of mixture models: Since PoEs multiply predictions made by independent GP experts, the overall prediction naturally weights the contribution of each expert. However, the model tends to be overconfident~\cite{Ng2014a}.
Cao and Fleet~\cite{Cao2014} recently proposed a generalised PoE-GP model in which the contribution of an expert in the overall prediction can weighted individually. This model is often too conservative, i.e., it over-estimates variances.
Tresp's Bayesian Committee Machine (BCM) \cite{Tresp2000} can be considered a PoE-GP model, which provides a consistent framework for combining independent estimators within the Bayesian framework, but it suffers from weak experts.

In this paper, we exploit the fact that the computations of PoE models can be distributed amongst individual computing units and propose the robust BCM (rBMC), a new family of hierarchical PoE-GP models that (i) includes the BCM~\cite{Tresp2000} and to some degree the generalised PoE-GP~\cite{Cao2014} as special cases, (ii) provides consistent approximations of a full GP, (iii) scales to arbitrarily large data sets by parallelisation. Unlike sparse GPs our rBCM operates on the full data set but distributes the computational and memory load amongst a large set of independent computational units. The rBCM recursively recombines these independent computations to form an efficient distributed GP inference\slash training framework.

A key advantage of the rBCM is that all computations can be performed analytically, i.e., no sampling is required. With sufficient computing power our model can handle arbitrarily large data sets. We demonstrate that the rBCM can be applied to data sets of size $\mathcal O(10^7)$, which exceeds the typical data set sizes sparse GPs deal with by orders of magnitude. However, even with limited resources, our model is practical: A GP with a million data points can be trained in less than half an hour on a laptop.

\section{Problem Set-up and Objective}
We consider a regression problem $y=f(\vec x)+\epsilon\in\R$, where $\vec
x\in\R^D$. The Gaussian likelihood $p(y|f(\vec x))=\gauss{f(\vec
  x)}{\sigma_\epsilon^2}$ accounts for the i.i.d. measurement noise
$\epsilon\sim\gauss{0}{\sigma_\epsilon^2}$. The objective is to infer
the latent function $f$ from a training data set $\mat X = \{\vec
x_i\}_{i=1}^N, \vec y = \{y_i\}_{i=1}^N$. For small data set
sizes $N$, a Gaussian process (GP) is a method of choice for
probabilistic non-parametric regression.
%
A GP is defined as a collection of random
variables, any finite number of which is Gaussian distributed. A GP is
fully specified by a mean function $m$ and a covariance function $k$
(kernel) with hyper-parameters $\vec\psi$. Without loss of generality, we assume
that the prior mean function is 0.

A GP is typically trained by finding hyper-parameters $\vec\theta =
\{\vec\psi, \sigma_\epsilon\}$ that maximise the log-marginal
likelihood
\begin{align}
  & \log p(\vec y|\mat X, \vec\theta) \nonumber\\
  & \stackrel{.}{=} -\tfrac{1}{2}\big(\vec y\T(\mat
  K+\sigma_\epsilon^2\mat I)\inv\vec y + \log|\mat
  K+\sigma_\varepsilon^2\mat I|\big)\,,
\label{eq:log-marginal likelihood}
\end{align}
where $\mat K=k(\mat X, \mat X)\in\R^{N\times N}$ is the kernel
matrix.  

For a given set of hyper-parameters $\vec\theta$, a training set $\mat
X, \vec y$ and a test input $\vec x_*\in\R^D$, the GP posterior
predictive distribution of the corresponding function value $f_* =
f(\vec x_*)$ is Gaussian with mean and variance given by
\begin{align}
  \E[f_*] &= m(\vec x_*) = \vec k_*\T(\mat K + \sigma_\varepsilon^2\mat I)\inv \vec y\,,\label{eq:mean GP}\\
  \var[f_*] &= \sigma^2(\vec x_*) = k_{**} -\vec k_*\T(\mat K +
  \sigma_\varepsilon^2\mat I)\inv \vec k_*\,,\label{eq:var GP}
\end{align}
respectively, where $\vec k_* = k(\mat X,\vec x_*)$ and
$k_{**}=k(\vec x_*, \vec x_*)$. 

Training requires the inversion and the determinant of \mbox{$\mat K +
  \sigma_\epsilon^2\mat I$} in~\eqref{eq:log-marginal likelihood},
both of which scale in $\mathcal{O}(N^3)$ with a standard
implementation.
For predictions, we cache $(\mat K + \sigma_\epsilon^2\mat I)\inv$,
such that the mean and variance in~\eqref{eq:mean GP} and
\eqref{eq:var GP} require $\mathcal{O}(N)$ and $\mathcal{O}(N^2)$
computations, respectively.
%
For $N>10,000$ training and predicting become time-consuming
procedures, which additionally require $\mathcal{O}(N^2+ND)$ memory.

Throughout this paper, we assume that a standard GP is a good model for
the latent function $f$. However, due to the data set size $N$ the
full GP is not applicable.

To scale GPs to large data sets with $N\gg \mathcal O(10^4)$, we address both the computational and the memory issues of full GPs by distributing the computational and memory loads to many independent computational units that only operate on subsets of the data. For this purpose, we devise a robust and scalable hierarchical product-of-GP-experts model.

\section{Distributed Product-of-GP-Experts Models}
\label{sec:hierarchical poe models}
\begin{figure*}[tb]
\centering
\subfigure[1-layer model.]{
\includegraphics[width = 0.31\hsize]{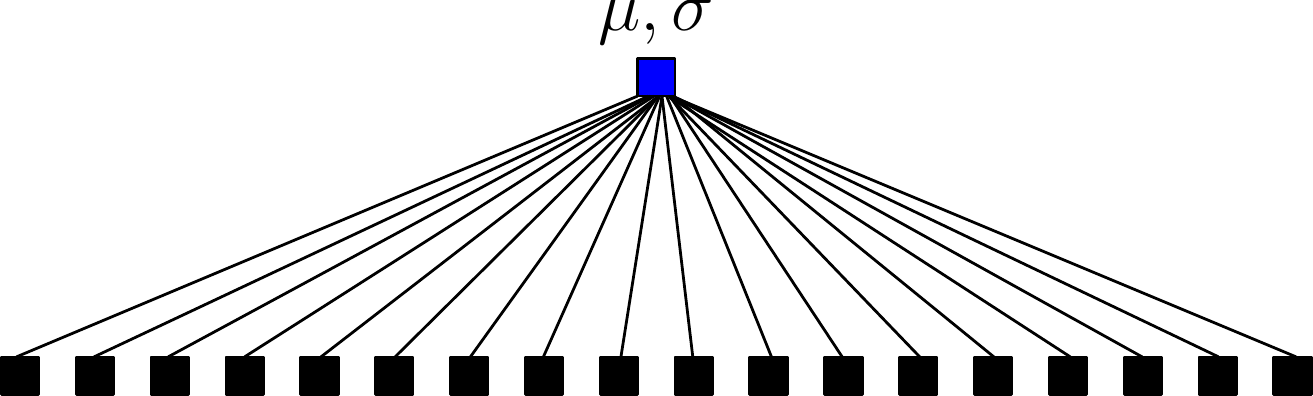}
\label{fig:1layer}
}
\hfill
\subfigure[2-layer model.]{
\includegraphics[width = 0.31\hsize]{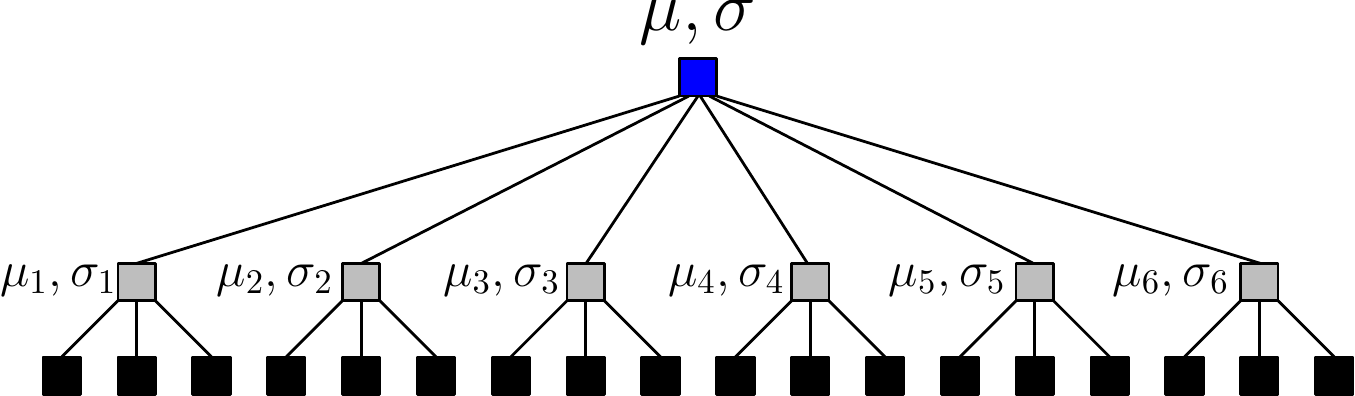}
\label{fig:2layer}
}
\hfill
\subfigure[3-layer model.]{
\includegraphics[width = 0.31\hsize]{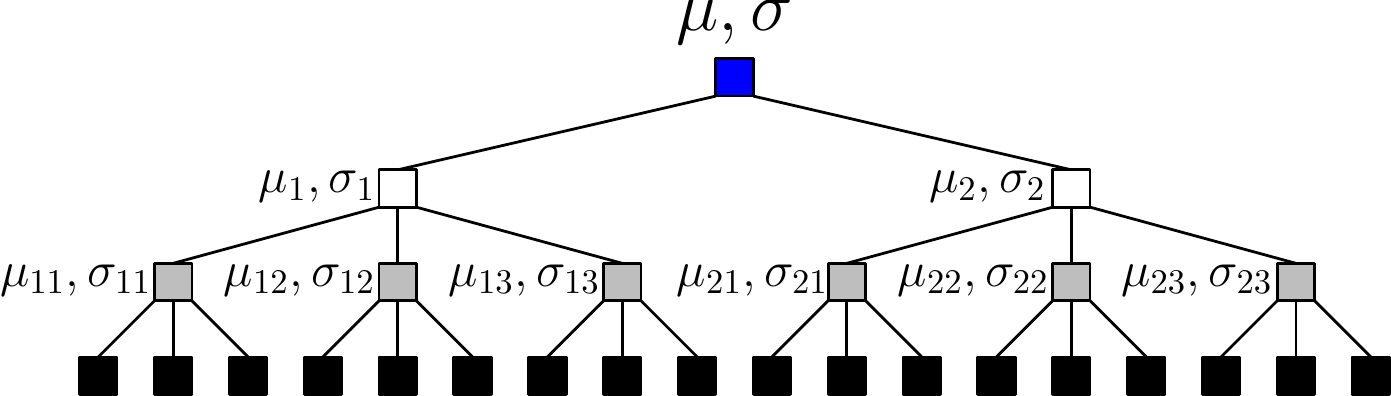}
\label{fig:3layer}
}
\caption{Computational graphs of hierarchical PoE models. Main computations are at the leaf nodes (GP experts, black). All other nodes recombine computations from their direct children. The top node (blue) computes the overall prediction.}
\label{fig:compGraphs}
\end{figure*}
Product-of-experts models (PoEs) are generally promising for parallelisation and distributed computing. In a PoE model, an overall computation is the product of many independent (smaller) computations, performed by ``experts''. In our case, every expert is a GP that accesses only a subset of the training data. 
In this paper, we consider a GP with a training data set $\mathcal{D} = \{\mat X, \vec y\}$. We partition the training data into $M$ sets $\mathcal D^\idx{k}=\{\mat X^\idx{k}, \vec y^\idx{k}\}$, $k=1,\dotsc,M$, and use a GP on each of them as a (local) expert\footnote{The notion of ``locality'' is misleading as our model does not require similarity measures induced by stationary kernels.}. Each GP expert performs computations (e.g., mean\slash variance predictions) conditioned on their respective training data $\mathcal D^\idx{k}$. These (local) predictions are recombined by a parent node (see Fig.~\ref{fig:1layer}), which subsequently may play the role of an expert at the next level of the model architecture. Recursive application of these recombinations results in a multi-layered tree-structured computational graph, see Fig.~\ref{fig:3layer}.

The assumption of independent GP experts leads to a block-diagonal approximation of the kernel matrix, which (i) allows for efficient training and predicting (ii) can be computed efficiently (time and memory) by parallelisation. 

\subsection{Training}
\label{sec:poe training}
Due to the independence assumption, the marginal likelihood $p(\vec y|\mat X, \vec\theta)$ in a PoE model factorises into the product of $M$ individual terms, such that
\begin{align}
p(\vec y|\mat X,\vec\theta) \approx \prod\nolimits_{k=1}^M p_k(\vec
y^\idx{k}|\mat X^\idx{k},\vec\theta)\,,
\end{align}
where each factor $p_k$ is determined by the $k$th GP expert. 
For training the PoE model, we seek GP hyper-parameters $\vec\theta$ that maximise the corresponding log-marginal likelihood
\begin{align}
\log p(\vec y| \mat X, \vec\theta) \approx \sum\nolimits_{k=1}^M \log
p_k(\vec y^\idx{k}|\mat X^\idx{k}, \vec\theta) 
\label{eq:approximate log-marginal likelihood}
\end{align}
where $M$ is the number of GP experts. The terms in~\eqref{eq:approximate log-marginal likelihood} are independently computed and given by
\begin{equation}
\label{eq:term in approx. LML}
\begin{aligned}
\hspace{-3mm}\log p(\vec y^\idx{k}|\mat X^\idx{k},\vec\theta) &= -\tfrac{1}{2}\vec y^\idx{k}(\mat K_{\vec\psi}^\idx{k} + \sigma_\epsilon^2\mat I)\inv\vec y^\idx{k}\\
&\quad -\tfrac{1}{2}\log |\mat K_{\vec\psi}^\idx{k} + \sigma_\epsilon^2\mat I| + \text{const}\,,
\end{aligned}
\end{equation}
%
where $\mat K_{\vec\psi}^\idx{k} = k(\mat X^\idx{k}, \mat X^\idx{k})$ is an $n_k\times n_k$ matrix, and $n_k\ll N$ is the size of the data set associated with the $k$th GP expert. 
Since we assume that a standard GP is sufficient to model the latent function, all GP experts at the leaves of the tree-structured model are trained jointly and share a single set of hyper-parameters $\vec\theta=\{\vec\psi,\sigma_\epsilon\}$.
Computing the log-marginal likelihood terms in~\eqref{eq:term in approx. LML} requires the inversion and determinant of $\mat K_{\vec\psi}^\idx{k} +\sigma_\epsilon^2\mat I$. These computations require $\mathcal{O}(n_k^3)$ time with a standard implementation. The memory consumption is $\mathcal{O}(n_k^2 + n_kD)$ for each GP expert. 

In~\eqref{eq:approximate log-marginal likelihood}, the number of parameters $\vec\theta$ to be optimised is relatively small since we do not consider additional variational parameters or inducing inputs that we optimise. Training (i) allows for straightforward parallelisation, (ii) provides a significant speed-up compared to training a full GP, (iii) requires solving low-dimensional $\mathcal O(D)$ optimisation problem unlike sparse GPs, which additionally optimise inducing inputs or variational parameters.

\subsection{Predictions}
\label{sec:pathologies}
\begin{figure*}
\centering
\subfigure[Product of GP Experts.]{
\includegraphics[width = 0.23\hsize]{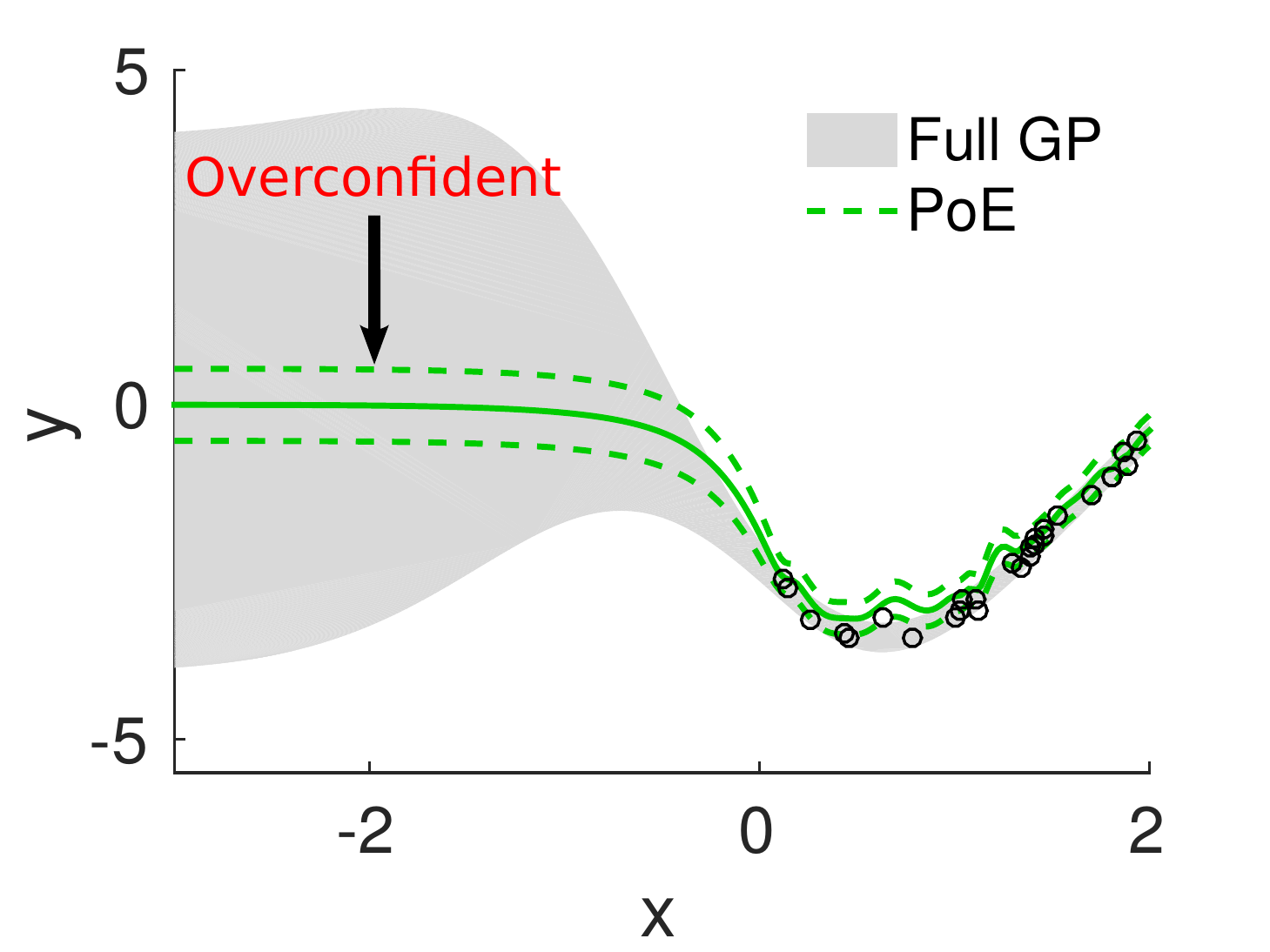}
\label{fig:HGP pathology}
}
\hfill
\subfigure[Generalised Product of GP Experts.]{
\includegraphics[width = 0.23\hsize]{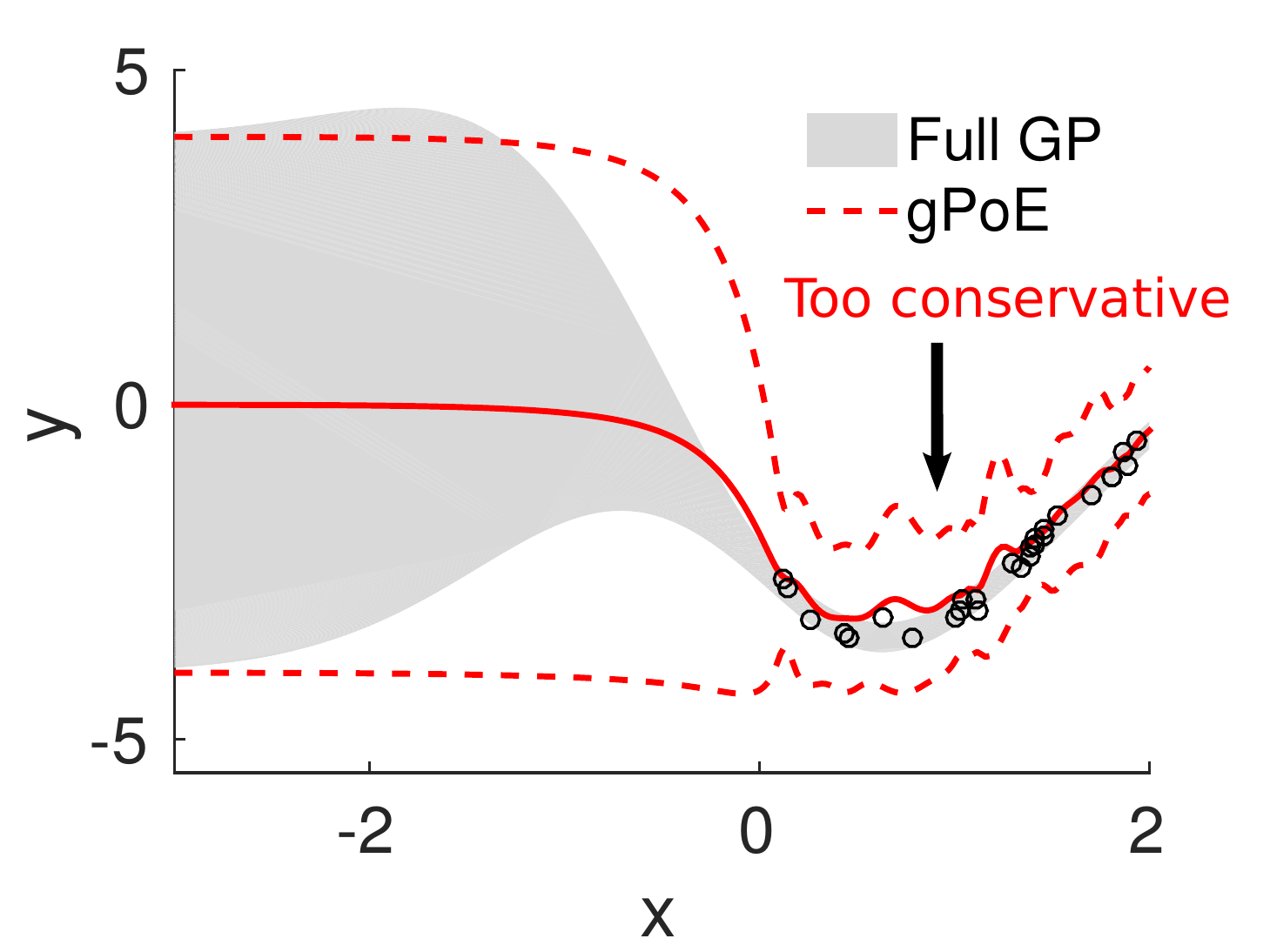}
\label{fig:gPoE pathology}
}
\hfill
\subfigure[Bayesian Committee Machine.]{
\includegraphics[width = 0.23\hsize]{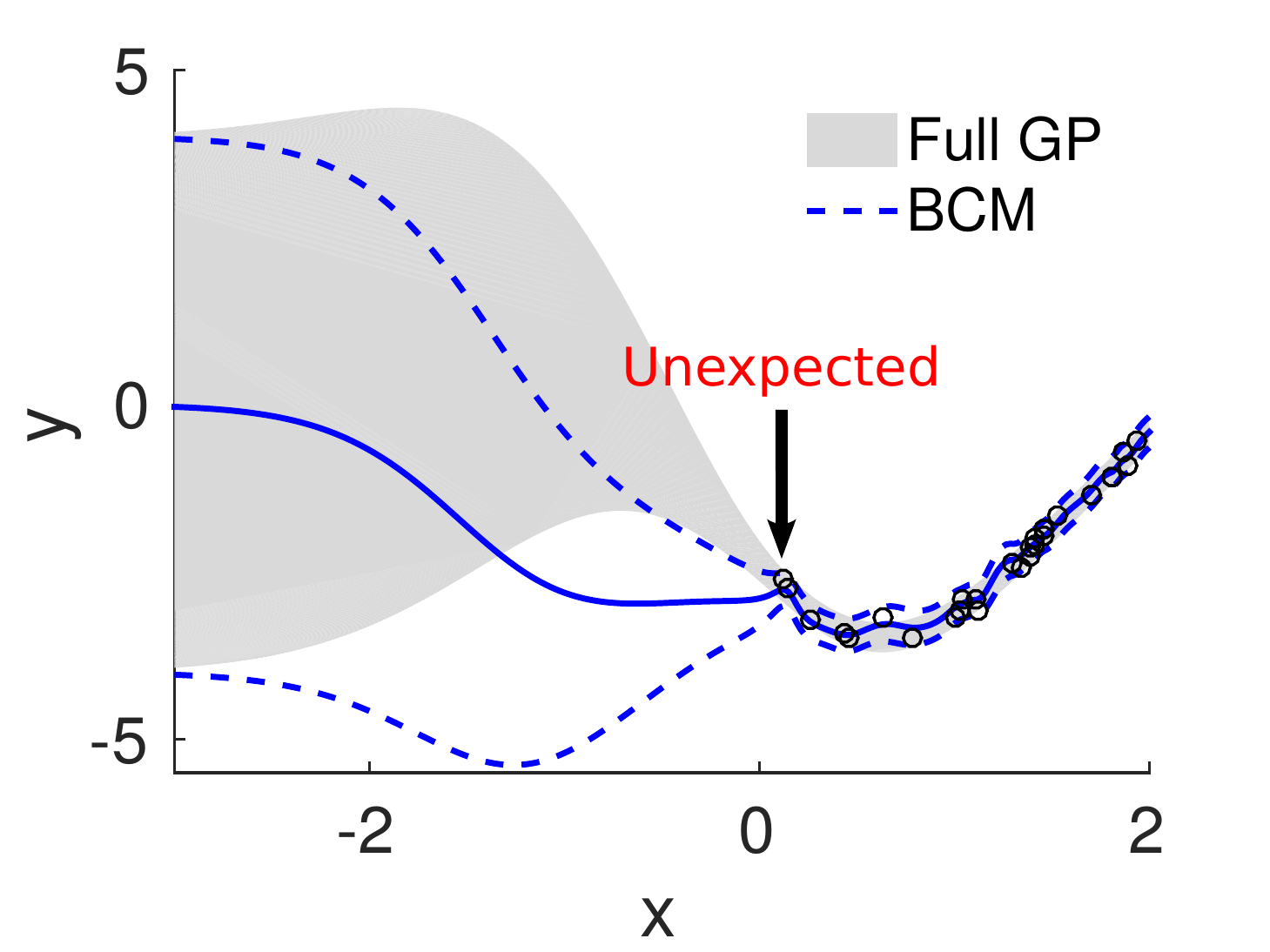}
\label{fig:BCM pathology}
}
\hfill
\subfigure[Robust Bayesian Committee Machine.]{
\includegraphics[width = 0.23\hsize]{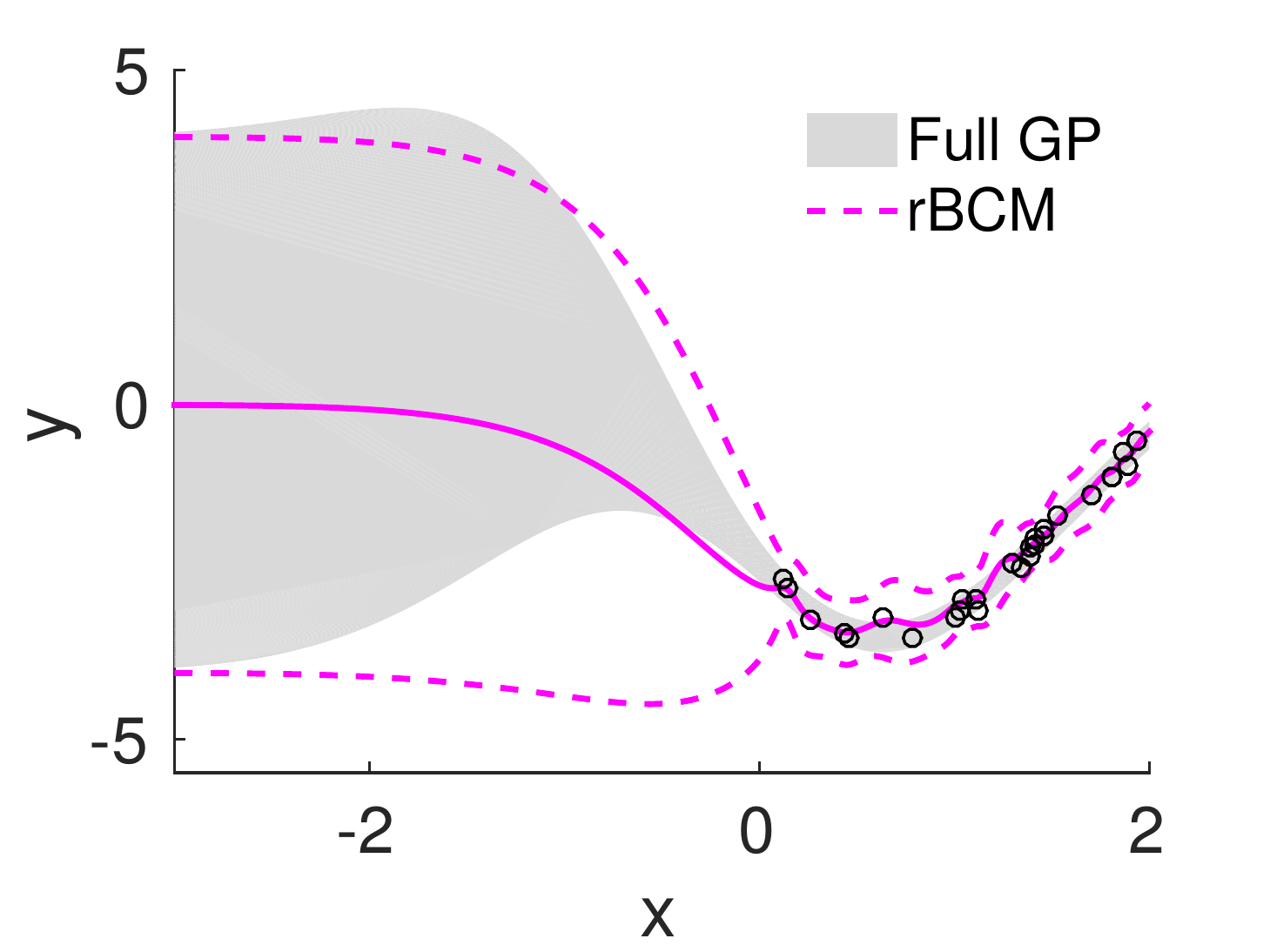}

\label{fig:rBCM}
}
\caption{Predictions with four different product-of-experts
  models. The full GP to be approximated is shown by the shaded area,
  representing 95\% of the GP confidence intervals. Training data is
  shown as black circles. Each GP expert was assigned two data
  points. \subref{fig:HGP pathology}: The standard PoE model does not
  fall back to the prior when leaving the training
  data. \subref{fig:gPoE pathology}: The generalised PoE model falls
  back to the prior, but over-estimates the variances in the regime of
  the training data. \subref{fig:BCM pathology}: The BCM is close to
  the full GP in the range of the training data, but the predictive
  mean can suffer. \subref{fig:rBCM}: The rBCM is more robust to weak experts than the (g)PoE and the BCM and produces reasonable predictions.}
\label{fig:pathologies}
\figspace
\end{figure*}
In the following, we assume that a set of $M$ GP experts has been trained according to Section~\ref{sec:poe training} and detail how the PoE~\cite{Ng2014a}, the generalised PoE~\cite{Cao2014} and the Bayesian Committee Machine~\cite{Tresp2000} combine predictions of the GP experts to form an overall prediction. Furthermore, we highlight strengths and weaknesses of these models, which motivates our robust Bayesian Committee Machine (rBCM). The rBCM unifies many other models while providing additional flexibility, which can address the shortcomings of the PoE, gPoE and the BCM. For illustration purposes, we focus on the model in Fig.~\ref{fig:1layer}, but many models generalise to an arbitrarily deep computational graph (see Section~\ref{sec:distributed computations}).

\subsubsection{Product of GP Experts}
\label{sec:poe}
The product-of-GP-experts model predicts a function value $f_*$ at a
corresponding test input $\vec x_*$ according to
\begin{align}
p(f_*|\vec x_*,\mathcal{D}) = \prod\nolimits_{k=1}^M p_k(f_*|\vec
x_*,\mathcal{D}^\idx{k})\,,
\label{eq:poe}
\end{align}
where $M$ GP experts operate on different training data subsets $\mathcal{D}^\idx{k}$. The $M$ GP experts predict means $\mu_k(\vec x_*)$ and variances $\sigma^2_k(\vec x_*)$, $k=1,\dotsc,M$, independently. The joint prediction $p(f_*|\vec x_*,\mathcal D)$ is obtained by the product of all experts' predictions. The product of these Gaussian predictions is proportional to a Gaussian with mean and precision
\begin{align}
\mu_*^{\text{poe}} &= (\sigma^\text{poe}_*)^2\sum\nolimits_k\sigma_k^{-2}(\vec x_*)\mu_k(\vec
x_*)\,,\label{eq:pred mean PoE}\\
(\sigma_*^{\text{poe}})^{-2} &= \sum\nolimits_k \sigma_k^{-2}(\vec x_*)\,,
\label{eq:pred var PoE}
\end{align}
respectively. For $k=1$, this model is identical to the full GP we wish to approximate.

Strengths of the PoE model are (a) the overall prediction $p(f_*|\vec x_*, \mathcal D)$ is straightforward to compute, (b) there are no free weight parameters to be assigned to each prediction (unlike in MoE models). A shortcoming of this model is that with an increasing number of GP experts the predictive variances vanish (the precisions add up, see~\eqref{eq:pred var PoE}), which leads to overconfident predictions, especially in regions without data. Thus, the PoE model is inconsistent in the sense that it does not fall back to the prior, see Fig.~\ref{fig:HGP pathology}.

\subsubsection{Generalised Product of GP Experts}
\label{sec:gpoe}
The generalised product-of-experts model (gPoE) by \citet{Cao2014}
adds the flexibility of increasing\slash reducing the importance of
 experts. The predictive distribution is
\begin{align}
\label{eq:gpoe model}
p(f_*|\vec x_*,\mathcal{D}) = \prod\nolimits_{k=1}^M p_k^{\beta_k}(f_*|\vec
x_*,\mathcal{D}^\idx{k})\,,
\end{align}
where the $\beta_k$ weight the contributions of the experts. The
predictive mean and precision are, therefore,
\begin{align}
\mu_*^\text{gpoe} &= (\sigma_*^\text{gpoe})^2\sum\nolimits_k \beta_k \sigma_k^{-2}(\vec
x_*)\mu_k(\vec x_*)\,,\label{eq:gpoe mean}\\
(\sigma_*^\text{gpoe})^{-2} &= \sum\nolimits_k \beta_k \sigma_k^{-2}(\vec x_*)\,,
\label{eq:gpoe precision}
\end{align}
respectively. 
A strength of the gPoE is that with $\sum_k\beta_k=1$ the model falls back to the prior outside the range of the data (and corresponds to a log-opinion-pool model~\cite{Heskes1998}). A weakness of the gPoE is that in the range of the data, it over-estimates the variances, i.e., the predictions are generally too conservative, especially with an increasing number of GP experts. Fig.~\ref{fig:gPoE pathology} illustrates these two properties.

\citet{Cao2014} suggest to set $\beta_k$ to the difference in the differential entropy between the prior and the posterior to determine the importance. In this paper, we do not consider this setting for the gPoE for two reasons: (i) $\sum_k\beta_k\neq 1$ leads to unreasonable error bars; (ii) Even with a normalisation, this setting does not allow for deep computational graphs. In fact, it only allows for a single-layer computational graph, such as Fig.~\ref{fig:1layer}. To allow for a general computational graph, we require an a-priori setting of the $\beta_k$. Thus, we set $\beta_k=1/M$, where $M$ is the number of GP experts. In this case, the predicted means in~\eqref{eq:pred mean PoE} and~\eqref{eq:gpoe mean} are identical, but the precisions differ, see also Fig.~\ref{fig:HGP pathology} and~\ref{fig:gPoE pathology} for a comparison.

\subsubsection{Bayesian Committee Machine}
\label{sec:bcm}
A third model that falls in the category of PoE models is the Bayesian
Committee Machine (BCM) proposed by \citet{Tresp2000}. Unlike the
(g)PoE, the BCM explicitly incorporates the GP prior $p(f)$ when
combining predictions (and not only at the leaves).

For two experts $j,k$ and corresponding training data sets $\mathcal
D^\idx{j}, \mathcal D^\idx{k}$, the predictive distribution is
generally given by
\begin{align}
  p(f_*| \mathcal D^\idx{j}, \mathcal D^\idx{k})\propto p(\mathcal
  D^\idx{j},\mathcal D^\idx{k}|f_*) p(f_*)\,,
  \label{eq:general prediction bcm derivation}
\end{align}
where $p(f_*)$ is the GP prior over functions. The BCM makes the conditional independence assumption that $\mathcal D^\idx{j}\perp\!\!\!\perp \mathcal D^\idx{k}|f_*$. With~\eqref{eq:general prediction bcm derivation} this yields 
\begin{align}
  p(f_*| \mathcal D^\idx{j}, \mathcal D^\idx{k}) 
  &\stackrel{\text{BCM}}{\propto} p(\mathcal
  D^\idx{k}|f_*) p(\mathcal D^\idx{j}|f_*)p(f_*)\label{eq:bcm
    derivation begin}\\
  &= \frac{ p(\mathcal
    D^\idx{k},f_*) p(\mathcal D^\idx{j},f_*)}{p(f_*)}\\
  &\propto \frac{p_k(f_*|\mathcal D^\idx{k}) p_j(f_*|\mathcal
    D^\idx{j})}{p(f_*)}\label{eq:bcm derivation end}\,,
\end{align}
which is the PoE model in~\eqref{eq:poe} divided by the GP prior. 
For $M$ training data sets $\mathcal D^\idx{k}$, $k=1,\dotsc,M$, the BCM applies the above approximation repeatedly, leading to the BCM's posterior predictive distribution
\begin{align}
p(f_*|\vec x_*, \mathcal{D}) &= \frac{\prod_{k=1}^M p_k(f_*|\vec x_*,
  \mathcal{D}^\idx{k})}{p^{M-1}(f_*|\vec x_*)} \,.
\label{eq:BCM model}
\end{align}
The ($M-1$)-fold division by the prior is the decisive difference between the BCM and the PoE model in~\eqref{eq:poe} and leads to the BCM's predictive mean and precision
\begin{align}
\mu_*^\text{bcm} &= (\sigma_*^\text{bcm})^2\sum\nolimits_{k=1}^M \sigma_k^{-2}(\vec
x_*)\mu_k(\vec x_*)\\
(\sigma_*^\text{bcm})^{-2} &= \sum\nolimits_{k=1}^M\sigma_k^{-2}(\vec x_*)  +
(1-M)\sigma_{**}^{-2}\,,
\label{eq:BCM precision}
\end{align}
respectively, where $\sigma_{**}^{-2}$ is the prior precision of
$p(f_*)$. 

The repeated application of Bayes's theorem leads to an
$(M-1)$-fold division by the prior in~\eqref{eq:BCM model}, which plays the role of a ``correction'' term in~\eqref{eq:BCM precision} that ensures a consistent model that falls back to the prior.\footnote{The BCM predictions fall into the category of normalised group odds~\cite{Bordley1982}.} The error bars of the BCM within the range of the data are usually good, but it is possible to ``break'' the BCM when only few data points are assigned to each GP expert. In Fig.~\ref{fig:BCM pathology}, we see that the posterior mean suffers from weak experts when leaving the data (around $x=0$).\footnote{Here, each expert was assigned only two data points.}

\subsubsection{Robust Bayesian Committee Machine}
\label{sec:gbcm}
In this section, we propose the robust Bayesian Committee Machine (rBCM), a unified model that (a) includes the gPoE and BCM as special cases, (b) yields consistent predictions, (c) can be implemented on a distributed computing architecture.

Inspired by the gPoE in~\eqref{eq:gpoe model}, the rBCM is a
BCM with the added flexibility of increasing\slash decreasing an expert's importance. The rBCM's predictive distribution is
\begin{align}
\label{eq:gbcm model}
  p(f_*|\vec x_*, \mathcal{D}) &= \frac{\prod\nolimits_{k=1}^M
    p_k^{\beta_k}(f_*|\vec x_*,
    \mathcal{D}^\idx{k})}{p^{-1+\sum_k\beta_k}(f_*|\vec x_*)}\,,
\end{align}
where the predictive mean and precision are given as
\begin{align}
  \mu_*^\text{rbcm} &= (\sigma^\text{rbcm}_*)^2\sum\nolimits_k
  \beta_k\sigma_k^{-2}(\vec x_*) \mu_k(\vec x_*)
\label{eq:rBCM mean}\\
(\sigma^\text{rbcm}_*)^{-2} &= \sum\nolimits_{k=1}^M \beta_k \sigma^{-2}_k(\vec x_*)
+ (1-\sum\nolimits_{k=1}^M \beta_k) \sigma_{**}^{-2}\,,
\label{eq:rBCM precision}
\end{align}
respectively.  The derivation of the rBCM in~\eqref{eq:gbcm model} is analogous to the BCM's derivation in~\eqref{eq:bcm derivation begin}--\eqref{eq:bcm derivation end}.
  
The rBCM combines the flexibility of the generalised PoE with the appropriate Bayesian treatment of the BCM, which leads to the correction term $(1-\sum\nolimits_k \beta_k) \sigma_{**}^{-2}$ in the the precision in~\eqref{eq:rBCM precision}. This correction term ensures that the predictive variance falls back to the prior when leaving the data. Note that we no longer require $\sum_k\beta_k = 1$ to ensure this (see also~\cite{Bordley1982}), which will facilitate computational graphs with multiple layers. 
The gPoE from Section~\ref{sec:gpoe} and the BCM from
Section~\ref{sec:bcm} are recovered for $\beta_k=1/M$ and $\beta_k=1$,
respectively. For $\beta_k=0$, the rBCM is identical to the GP prior,
and for $\beta_k=1$ but without the correction, the rBCM recovers the
PoE from Section~\ref{sec:poe}

The parameters $\beta_k$ control not only the importance of the individual experts, but they also control how strong the influence of the prior is. Assuming each GP expert is a good predictive model, we would set $\beta_k=1$ for all $k$, such that we retain the BCM. If the quality of the GP experts is weak, e.g., data is noisy and the experts' data sets $\mathcal{D}^\idx{k}$ are small, $\beta_k$ allows us to weaken the experts' votes and to robustify the predictive model by putting (relatively) more weight on the prior. Therefore, we follow the suggestion by \citet{Cao2014} and choose $\beta_k$ according to the predictive power of each expert at $\vec x_*$. Specifically, we use the difference in differential entropy between the prior $p(f_*|\vec x_*)$ and the posterior $p_k(f_*|\vec x_*, \mathcal D^\idx{k})$. This quantity can be computed efficiently and is given as $\beta_k=\tfrac{1}{2}(\log\sigma^2_{**} - \log\sigma_k^2(\vec x_*))$, where $\sigma_{**}^2$ is the prior variance and $\sigma_k^2(\vec x_*)$ is the predictive variance of the $k$th expert.
%

Fig.~\ref{fig:rBCM} shows that for this choice of $\beta_k$ the rBCM expresses more uncertainty about the learned model than the BCM: Due to the adaptive influence of the prior in~\eqref{eq:rBCM mean}--\eqref{eq:rBCM precision}, the variances within the range of the data (black circles) are slightly larger, but the predictive mean no longer suffers from the dominant ``kink'' at around $x=0$ compared to the BCM in Fig.~\ref{fig:BCM pathology}. Overall, the rBCM provides more reasonable predictions than any other model in Fig.~\ref{fig:pathologies}.

\section{Distributed Computations }
\label{sec:distributed computations}

\begin{wrapfigure}{r}{0.27\hsize}
\vspace{-5mm}
\centering
\includegraphics[width = \hsize]{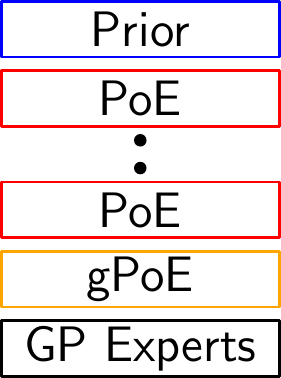}
\caption{General architecture for the rBCM with arbitrarily many
  layers.}
\label{fig:general architecture}
\vspace{-3mm}
\end{wrapfigure}
In the following, we show that for a given number $M$ of GP experts, the rBCM can be implemented in different computational graphs while providing identical predictions. For instance, with 32 experts, we show that a single-layer computational graph with 32 experts and one central node, see Fig.~\ref{fig:1layer}, is equivalent to a two-layer computational graph with 32 experts, 8 parent nodes (each of which is responsible for 4 GP experts), and one central node. This property can be exploited in distributed systems or to balance the communication between computing units.

In a single-layer model as shown in Fig.~\ref{fig:1layer} the rBCM predictions in~\eqref{eq:gbcm model} can be constructed by a gPoE (numerator) combining predictions of GP experts, followed by a correction via the prior (denominator). Let us consider a two-layer model  with $M$ GP experts, $L$ nodes at level 1 and one central node at level 2, see Fig.~\ref{fig:2layer}. Each grey node at level 1 is responsible for $L_k$ GP experts (black nodes). All $M$ GP experts $k_i$, compute weighted predictive distributions $p_{k_i}^{\beta_{k_i}}(f_*|\mathcal{D}^\idx{k_i})$. These predictions are then multiplied by the corresponding $L$ parent nodes (grey) to $p_k^{\beta_k}(f_*|\mathcal{D}^\idx{k})$, where $\beta_k = \sum_i\beta_{k_i}$ is the overall weight of the subtree following the $k$th node $L_k$ and $\mathcal D^\idx{k}=\bigcup_i \mathcal D^\idx{k_i}$. The overall prediction at the top node (blue in Fig.~\ref{fig:compGraphs}) is
\begin{align}
  \hspace{-2mm} \frac{\prod_{k=1}^M
    p_k^{\beta_k}(f_*|\mathcal{D}^\idx{k})}{\color{blue}{p^{\sum_k\beta_k
        -1}(f_*)}} =
  \frac{{\color{red}{\prod_{k=1}^L}}\prod_{i=1}^{L_k}p_{k_i}^{\beta_{k_i}}(f_*|\mathcal{D}^\idx{k_i})}{\color{blue}{p^{\sum_k\beta_k
        -1}(f_*)}}
\label{eq:gbcm 2-layer equivalence}
\end{align}
where we accounted for the $(\sum_k\beta_k -1)$-fold correction by the prior. This computation can be obtained by a gPoE (black), followed by a PoE (red) and a prior correction (blue) in~\eqref{eq:gbcm 2-layer equivalence}. Hence, for a given number of GP experts, the rBCM predictions can be equivalently realised in a single and two-layer computational graph, see Fig.~\ref{fig:compGraphs}.

This can be generalised further to an arbitrarily deep computational graph, whose general implementation structure is shown in Fig.~\ref{fig:general architecture}. The GP experts at the leaves compute their individual means, variances and confidence values $\beta_i$. The next layer consists of gPoE models, which compute the weighted means and variances according to~\eqref{eq:pred mean PoE} and~\eqref{eq:gpoe precision}, respectively (plus their overall weights $\beta_k=\sum_{i\in\text{children}}\beta_{k_i}$, which are passed on to the next-higher level). The gPoE is followed by an arbitrary number of PoE models, which compute means and precisions according to~\eqref{eq:pred mean PoE} and~\eqref{eq:pred var PoE}, respectively (plus their overall weights $\beta_k=\sum_{j\in\text{children}}\beta_{k_j}$, which are passed on to the next-higher level). The top layer accounts for the prior (blue term in~\eqref{eq:gbcm 2-layer equivalence}), which uses all the $\beta_k$ from the subtrees starting at its children to compute the overall mean and precision according to~\eqref{eq:rBCM mean} and~\eqref{eq:rBCM precision}, respectively.

Hence, for a given number of GP experts, there are many equivalent computational graphs for the rBCM (and the other distributed GPs discussed in Section~\ref{sec:hierarchical poe models}). For instance, all computational graphs in Fig.~\ref{fig:compGraphs} return identical predictive distributions. This allows us to choose the computational graph that works best with the computing infrastructure available: Shallow graphs minimise overall traffic. However, they are more vulnerable to communication bottlenecks at the central node since it has a large number of connections. Deeper computational graphs cause more overall communication, but the tree has a smaller branching factor. In practice, for a set of computational graphs we took the time it requires to compute the gradient of the marginal likelihood and chose the ``fastest'' architecture.

\section{Experiments}
We empirically assess three aspects of all distribute GP models: (1) The required training time, (2) the approximation quality, (3) a comparison with state-of-the-art sparse GP methods. 
In all experiments, we chose the standard squared exponential kernel with automatic relevance determination and a Gaussian likelihood.
%
Moreover, we assigned training data to experts \emph{randomly} for two reasons: First, we demonstrate that our models do not need locality information; second, random assignment is very fast compared to
clustering methods, e.g., KD-trees.

\subsection{Training Time for Large Data Sets}
\begin{figure}
\centering
\includegraphics[width = 0.8\hsize]{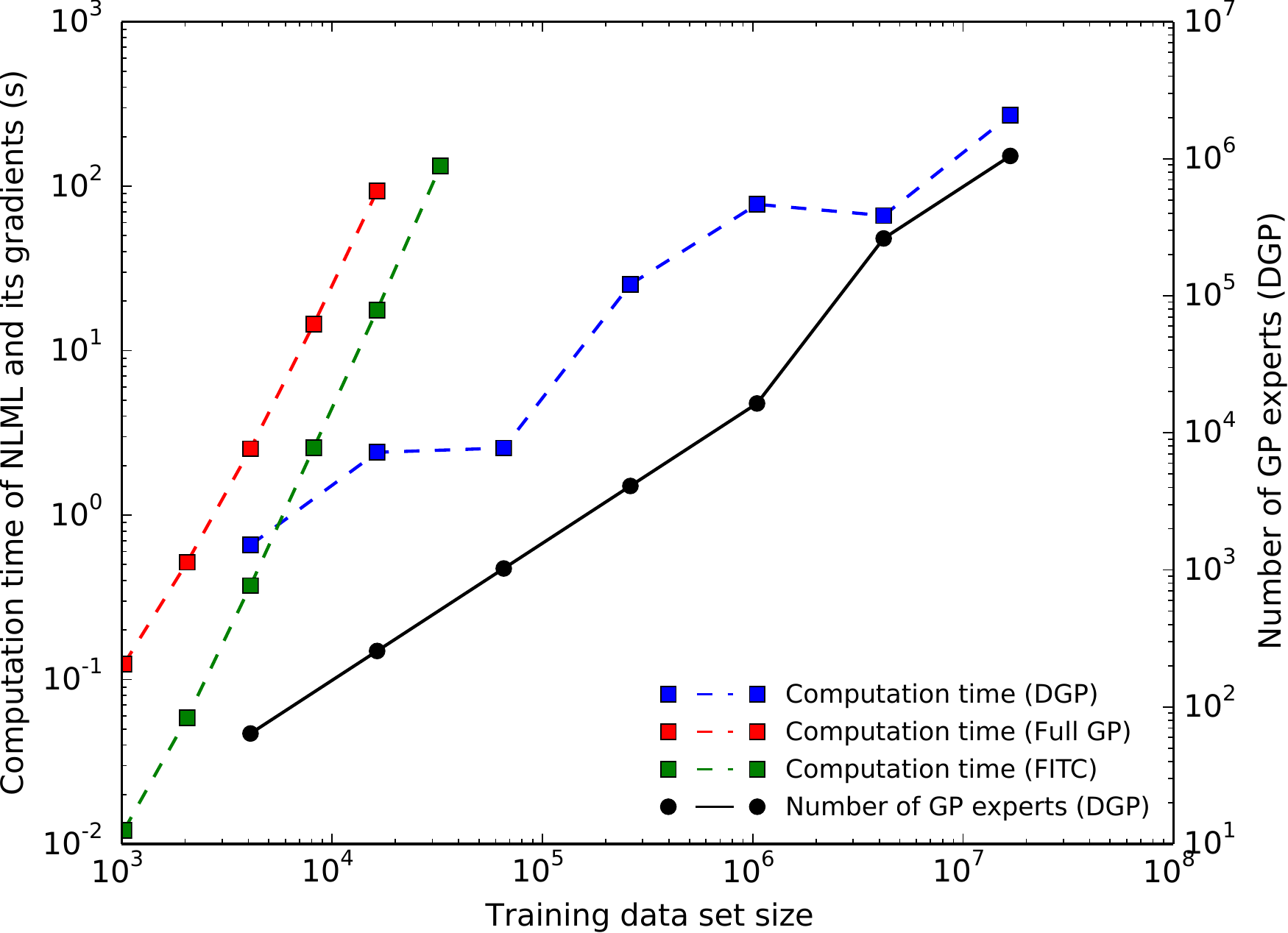}
\caption{Computation time for the log-marginal likelihood and its gradient with respect to the kernel hyper-parameters as a function of the size of the training data. The distributed GPs (DGP), i.e., (g)PoE and (r)BCM, scale favourably to large-scale data sets.}
\label{fig:hgp timings}
\figspace
\end{figure}
To evaluate the training time for distributed GPs (DGP)\footnote{This comprises the (g)PoE and (r)BCM for which the training time is identical.}, we measured the time required to compute the log-marginal likelihood and its gradient with respect to the kernel hyper-parameters. Since the model is trained using LBFGS, the overall training time is proportional to the time it takes to compute the log-marginal likelihood and its gradient.
For this evaluation, we chose a computer architecture of 64 nodes with four cores each. Furthermore, we chose a three-layer computational graph with varying branching factors. For data sets of $\leq 2^{20}$ data points each GP expert possessed 512 data points, for data set sizes of $>2^{20}$, we chose the number of data points per node to be 128.

Fig.~\ref{fig:hgp timings} shows the time required for computing the,log-marginal likelihood and its gradient with respect to the hyper-parameters. 
The horizontal axis shows the size of the training set (logarithmic scale), the left vertical axis shows the computation time in seconds (logarithmic scale) for the DGP (blue-dashed), a full GP (red-dashed) and a sparse GP (FITC) with inducing inputs~\cite{Snelson2006} (green-dashed). For the sparse GP model, we
chose the number $M$ of inducing inputs to be 10\% of the size of the training set, i.e., the computation time is of the order of $\mathcal{O}(NM^2)=\mathcal{O}(N^3/100)$, which offsets the curve of the full GP. The right vertical axis shows the number of GP experts (black-solid) amongst which we distribute the
computation.  
While the training time of the full GP becomes impractical at data set sizes of about 20,000, the sparse GP model can be reasonably trained up to 50,000 data points.\footnote{We did not include any computational overhead for learning the inducing inputs, which is often time consuming.}  The computational time required for the DGP to compute the marginal likelihood and gradients is significantly lower than that of the full GP, and we scaled it up to $2^{24} \approx 1.7 \times 10^7$ training data points, which required about the same amount of time ($\approx 230$\,s) for training a full GP with $2^{14}\approx 1.6\times 10^4$ and a sparse GP with $2^{15}\approx 3.2\times 10^4$ data points. The figure shows that for any problem size we can find a computational graph that allows us to train the model within a reasonable amount of time.

Even if a big computing infrastructure is not available our model is useful in practice: We performed a full training cycle of the DGP with $10^6$ data points on a standard laptop in about 20 minutes. This
is also a clear indicator that the memory consumption of the DGP is relatively small.

\subsection{Empirical Approximation Errors}

We evaluate the predictive quality of the DGP models introduced in Section~\ref{sec:hierarchical poe models} on the Kin40K data set. The data set represents the forward dynamics of an 8-link all-revolute robot arm. The goal is to predict the distance of the end-effector from a target, given the joint angles of the eight links as features.
The Kin40K data set consists of 10,000 training points and 30,000 test points. We use the same split into training and test data as \citet{Seeger2003}, \citet{Lazaro-Gredilla2010}, and \citet{Nguyen2014}.

We considered two baselines: a full (ground truth) GP and the subset-of-data (SOD) approximation, which uses a random subset of the full training data set to train a sparse GP. Taking training time into account, \citet{Chalupka2013} identified the SOD method as a good and robust sparse GP approximation.  All approximate models (PoE, gPoE, BCM, rBCM, SOD) used the hyper-parameters from the full GP to eliminate issues with local optima.
For every model, we took the time for computing the gradient of the
marginal likelihood (training is proportional to this amount of
time). We selected the number of data points for SOD, such that the gradient computation time approximately matches the one of the DGPs.

Experiments were repeated 10 times to average out the effect of the random assignment of data points to experts and the selection of the subset of training data for the SOD approximation. For this experiment, we used a Virtual Machine with 16 3\,GHz cores and 8 GB RAM.

\begin{figure}
\centering
\subfigure[RMSE.]{
\includegraphics[width = 0.47\hsize]{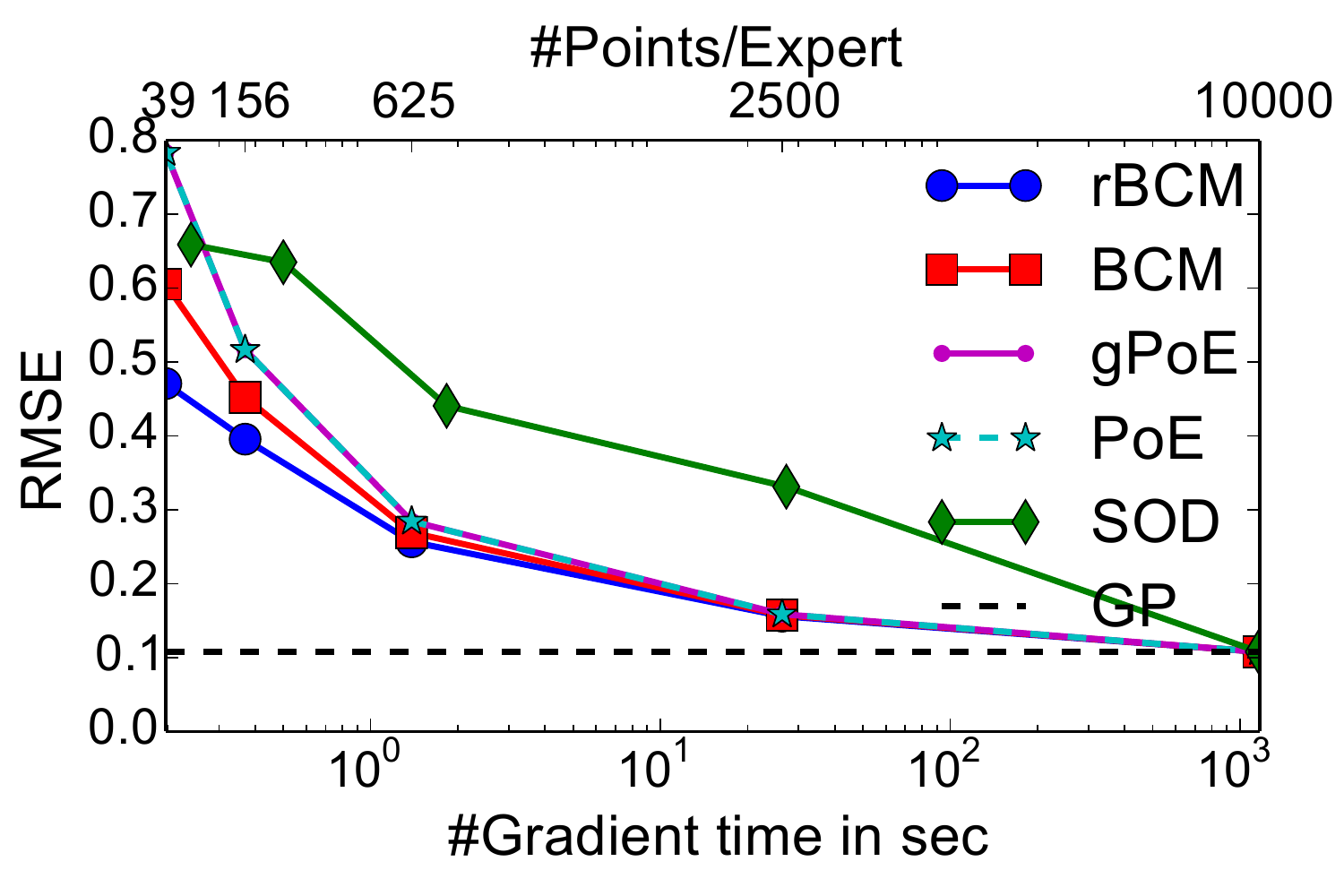}
\label{fig:time vs rmse}
}
\subfigure[NLPD.]{
\includegraphics[width = 0.47\hsize]{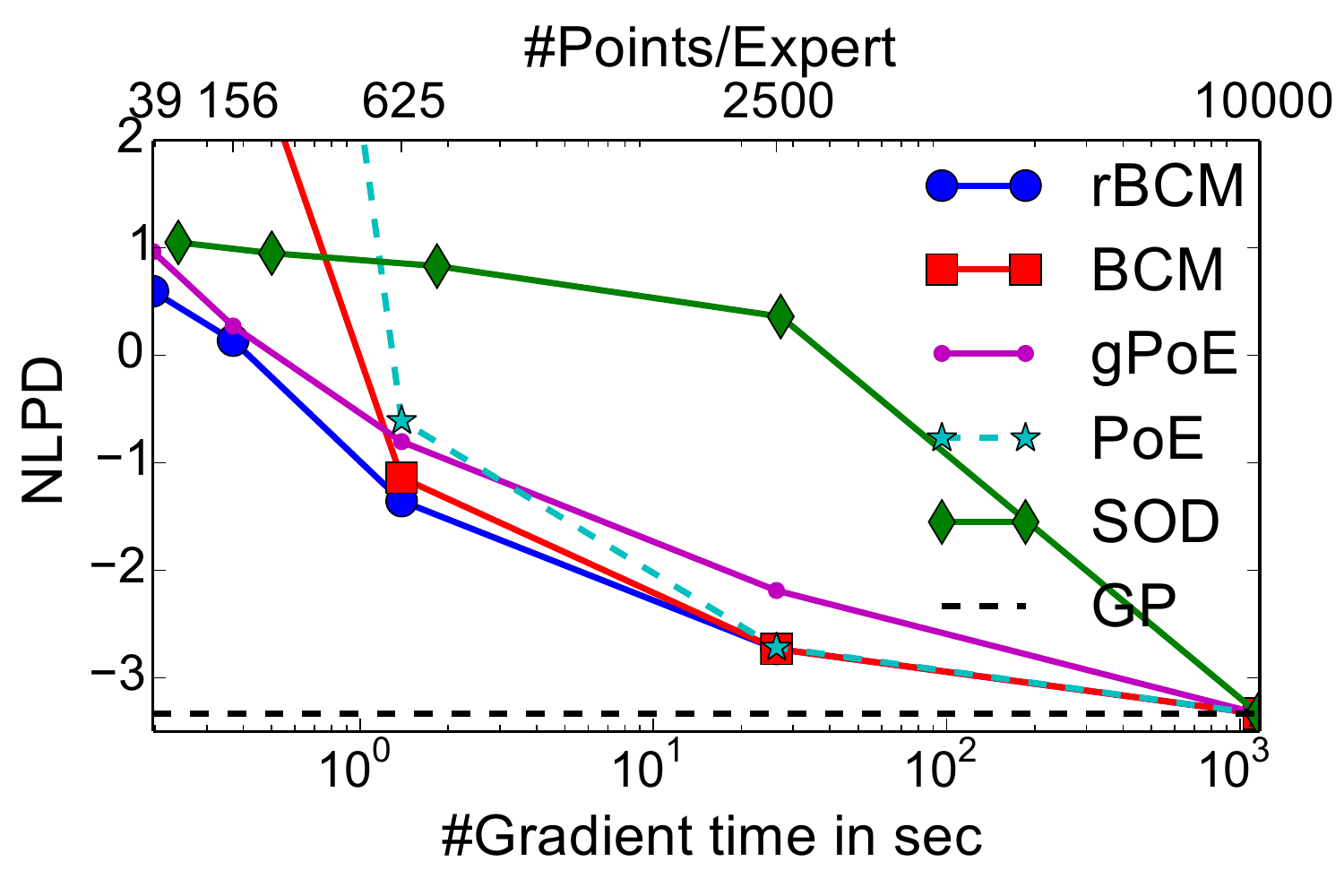}
\label{fig:time vs nlpd}
}
\caption{Performance as a function of training time (lower horizontal axes) and the number of data points per expert (upper horizontal axes).  \subref{fig:time vs rmse} RMSE,
  \subref{fig:time vs nlpd} NLPD. Standard errors (not displayed)
  are smaller than $10^{-2}$. }
\label{fig:kin40k times vs performance}
\figspace
\end{figure}
Fig.~\ref{fig:kin40k times vs performance} shows the average performance (RMSE and negative log predictive density (NLPD) per data point) of all models as a function of (a) the time to compute the gradient of the marginal likelihood and (b) the number of training data points per GP expert. The full GP (black, dashed) shows the ground-truth performance, but requires 1162 seconds for the gradient computation. The performances of the (r)BCM and the (g)POE have been evaluated using 256, 64, 16, 4, and 1 expert with 39, 156, 625, 2500 and 10000 data points per expert, respectively. In Fig.~\ref{fig:time vs rmse}, the rBCM consistently outperforms all other methods, where SOD is substantially worse than all DGPs. The NLPD in Fig.~\ref{fig:time vs nlpd} allows us to make some more conclusive statements: While the rBCM again outperforms all other methods, the BCM and the PoE's performances suffer when only a small number of data points is assigned to the GP experts. The PoE suffers from variance underestimation (see Fig.~\ref{fig:HGP pathology}) whereas the BCM suffers from ``weak'' experts (see Fig.~\ref{fig:BCM pathology}). SOD does not work well, even with 2500 data points. Overall, the rBCM provides an enormous training speed-up compared to a full GP, with a significantly better predictive performance than SOD.

\subsection{Airline Delays (US Flight Data)}
We assess the performance of the distributed GPs (PoE, gPoE, BCM and rBCM) on a large-scale non-stationary data set reporting flight arrival and departure times for every commercial fight in the US from January to April 2008\footnote{\url{http://stat-computing.org/dataexpo/2009/}}. This data set contains information about almost 6 million flights. We followed the procedure described by \citet{Hensman2013}\footnote{Thanks to J Hensman for the pre-processing script.} to predict the flight delay (in minutes) at arrival: We selected the first $P$ data points to train the model and the following 100,000 to test it. We chose the same eight input variables $\vec x$ as \citet{Hensman2013}: age of the aircraft, distance that needs to be covered, airtime, departure and arrival times, day of the week and month, month. This data set has been evaluated by~\citet{Hensman2013} and \citet{Gal2014}, who use either Stochastic Variational Inference GP (SVIGP) or Distributed Variational GPs (Dist-VGP) to deal with this training set size. 

Using a workstation with 12 3.5\,GHz cores and 32 GB RAM, we conducted 10 experiments with $P=7\times 10^5$, $P=2\times 10^6$ and $P=5\times 10^6$ where we chose 4096, 8192 and 32768 experts corresponding to 170, 244 and 152 training data points per expert, respectively. The computation of the marginal likelihoods required 13, 39, and 90 seconds, respectively.\footnote{All experiments can be conducted on a MacBook Air (2012) with 8\,GB RAM.} The computational graphs were (16-16-16), (8192), and (32-32-32), respectively, where each number denotes the branching factor at the corresponding level in the tree.
After normalising the data every single experiment consisted of an independent full training\slash test cycle, with random assignments of data points to GP experts. Training the DGPs normally required 30--100 line searches to converge.
\begin{table}[t]
\caption{US Flight Data Set. SVIGP and Dist-VGP results are reported from the
  respective papers. Results with $^+$ are the best solution
  found. Good and bad performances are highlighted in blue and red, respectively.}
\label{tab:flights}
\centering
\scalebox{0.8}{
\begin{tabular}{l|cc||cc||cc}
  & \multicolumn{2}{c}{700K/100K} &
  \multicolumn{2}{c}{2M/100K } &
  \multicolumn{2}{c}{5M/100K}\\
  \hline
  & RMSE & NLPD  & RMSE & NLPD   & RMSE & NLPD \\
  \hline
  SVIGP& $\red{33.0}$ & --- & --- & --- & --- & ---\\
  Dist-VGP$^+$& $\red{33.0}$ & --- & $35.3$  & --- & --- & --- \\
  rBCM  & $\blue{27.1}$ & $9.1$  & $\blue{34.4}$ & $\blue{8.4}$& $\blue{35.5}$ & $\blue{8.8}$\\
  BCM  & $\red{33.5 }$ & $\red{14.7} $  & $\red{55.8}$ & $\red{17.0} $ &$\red{55.3}$ & $\red{19.5}$ \\
  gPoE  & $28.7$ & $\blue{8.1}$  & $35.5$ & $\blue{8.6}$  &$37.3$ & $\blue{8.7}$\\
  PoE & $28.7$ & $\red{14.1}$   & $35.5$ & $\red{26.3}$ &$37.3$ & $\red{17.7}$\\
  SOD & $\red{>10^3}$ & $\red{22.6}$ & $\red{>10^3}$ & $\red{17.2}$ & $\red{>10^3}$ & $\red{16.4}$
\end{tabular}
}
\figspace
\end{table}
Table~\ref{tab:flights} reports the performance (RMSE and NLPD) of various large-scale GP methods for the flight data set.  The results for SVIGP and Dist-VGP are taken from \citet{Hensman2013} and \citet{Gal2014}, respectively.
Since SVIGP and Dist-VGP are difficult to optimise due to the large number of variational parameters, \citet{Gal2014} report only their best results, whereas we report an average of \emph{all} experiments conducted. 

The standard errors (not shown in Table~\ref{tab:flights}) of the rBCM and gPoE are consistently below 0.3, whereas the BCM and the PoE suffered from a few outliers, which is also indicated by the relatively large NLPD values. Compared to the Dist-VGP and SVIGP on the 700K data set, the rBCM, gPoE and PoE perform significantly better in RMSE. The table highlights the weaknesses of the PoE (under-estimation of the variance) and the BCM (problems with weak experts) very clearly. The property of the gPoE (too conservative) is a bit hidden: Although the RMSE of the gPoE is consistently worse than that of the rBCM, its NLPD tends to be a bit lower. The NLPD values of the rBCM and the gPoE are fairly consistent across all three experiments.

The data set exhibits the property that the 700K/100K data set is more stationary than the 2M/100K and 5M/100K data sets. Therefore, we observe a decreasing performance although we include more training data. This effect has already been reported by \citet{Gal2014}.

\section{Discussion and Conclusion}
The distributed GP models discussed in this paper exhibit the appealing property that they are conceptually simple: They split the data set into small pieces, train GP experts jointly, and subsequently combine individual predictions to an overall computation. Compared to sparse GP models, which are based on inducing or variational parameters, the distributed GPs possess only the normal GP hyper-parameters, i.e., it is less likely to end up in local optima. 

In this paper, all experts share the same hyper-parameters, which leads to automatic regularisation: The overall gradient is an average of the experts' marginal likelihood gradients, i.e., overfitting of individual experts is not favoured.

We believe that distributed computations are promising to scale GPs to truly large data sets. Sparse methods using inducing inputs are naturally limited by the number of inducing variables. In practice, we see $\mathcal O(10^2)$ many inducing variables (optimisation is hard), although their theoretical limit might be at $\mathcal O(10^4)$, if we assume that this is the limit up to which we can train a full GP. If we assume that a single inducing variable summarises 100 data points, the practical limit of sparse methods are data sets of size $\mathcal O(10^6)$. This can be extended by either a higher compression rate or parallelisation~\cite{Hensman2013, Gal2014}.

We introduced the robust Bayesian Committee Machine (rBCM), a conceptually straightforward, but effective, product-of-GP-experts model that scales GPs to (in principle) arbitrarily large data sets. The rBCM distributes computations amongst independent computing units, allowing for straightforward parallelisation. Recursive and closed-form recombinations of these independent computations yield a practical model that is both computationally and memory efficient.  The rBCM addresses shortcomings of other distributed models by appropriately incorporating the GP prior when combining predictions.  The rBCM is independent of the computational graph and can be equivalently used on heterogeneous computing infrastructures, ranging from laptops to large clusters. Training an rBCM with a million data points takes less than 30 minutes on a laptop.  With more computing power training the rBCM with $10^7$ data points can be done in a few hours.
Compared to state-of-the-art sparse GPs, our model is conceptually simple, performs well, learns fast, requires little memory and does not require high-dimensional optimisation.

\subsection*{Acknowledgements}
MPD has been supported by an Imperial College Junior Research
Fellowship.


\end{document}


\twocolumn[ \icmltitle{Distributed Gaussian Processes (Appendix)}]

\appendix

\section{Implementation Details}

In the following, we provide a few implementation details that are important for our python implementation. 

\subsection{True Concurrency in Python}
A known issue of the CPython interpreter, which we use in our
implementation, is the lack of true concurrency using the in-built
threading library. Due to the \emph{Global Interpreter Lock} (GIL,
which is implemented in the interpreter because Python's memory
management is not thread safe), only a single thread of Python code
can be executed at any point in time. Therefore, the use of threads in
the Python context only provides logical concurrency in terms of the
flow of programs, but not true simultaneous computations.

There exists a workaround for the true concurrency problem in Python,
via the use of processes instead of threads to perform simultaneous
computations. In the POSIX model, threads are lightweight units of
computations belonging to the same process, thus, sharing the same
memory space. Processes have their own memory space and come with
increased system overheads compared to threads. However, on Linux
(which we use for this implementation), the creation of duplicate
processes (forking) does not incur large memory costs since Linux
implements a copy-on-write model. This means that when a process forks
into two, the memory is not copied, unless the new process attempts to
modify it. In the context of our implementation, we make no
modification to the training data, which is shared amongst all
child-GPs. In terms of the memory  usage, each child-GP only needs to
compute its own kernel matrix and the corresponding Jacobian matrix
per hyper-parameter, which have no interaction with any other
child-GP. Therefore, computing each child-GP using a separate process
does not incur any large, redundant memory costs that would not be
present in a true concurrency model implemented by native threads.

\subsection{Memory Management}
This duplication of memory can be prevented, since the main training
data set is shared and does not get modified. Therefore, each GP expert
can have access to its training data subset without copying any of it.
To do this, we implement a DataSet class, which manages the
data. There will be a single instance of this class, which holds all
the data in the full training set. We can create additional DataSet
instances by invoking the subset method on the DataSet object. We
specify a set of integers corresponding to the indices of the data
points (which we require in the subset) in the main data set. A new
DataSet instance is then created with no actual data, but a list of
indices, and a reference to the main DataSet object as its
superset. The only exception to this occurs when distributed computing
is used, in which case, the subset of data that is required at a
different machine on the network is copied from one memory space to
another, and a new `main' DataSet object is created.

\subsection{Remote Object Management}
Managing network communications for a distributed system poses a large
challenge since there are many details one has to manage (e.g.,
retrying failed message transmissions, timeouts). This adds much
complexity to the system and may cause unnecessary failures in the
system if not properly implemented. The remote procedure call (RPC)
protocol enables all of the issues at the network layer to be
abstracted, and allows us to use objects on a remote host with the
same interface as local objects. This is done by having a dummy object
on the client (`client stub'), which provides the interface of the
object. However, instead of executing methods on the client, it
invokes the required computation on a remote host (`server'), and
(upon completion of the remote computation) receives the results from
the remote host and returns. With this interface, the code generally
remains simple and readable.  In our implementation, we use the Pyro4
library, which implements remote procedural calls in Python.  RPC is
used for communication between nodes in the gBCM tree. Furthermore,
the classes in our object-oriented implementation can be set up as
remote objects (as required).  This allows the exact functionality of
all classes for both local and distributed computation.